\documentclass[conference]{IEEEtran}
\IEEEoverridecommandlockouts
\usepackage{cite}
\usepackage{alltt}
\usepackage{amsmath,amssymb,amsfonts}
\usepackage{algorithmic}
\usepackage{graphicx}
\usepackage{textcomp}
\usepackage{xcolor}
\usepackage{hyperref}

\def\BibTeX{{\rm B\kern-.05em{\sc i\kern-.025em b}\kern-.08em
    T\kern-.1667em\lower.7ex\hbox{E}\kern-.125emX}}
\begin{document}

\title{Semi-automated checking for regulatory compliance in e-Health}

\author{\IEEEauthorblockN{1\textsuperscript{st} Ilaria Angela Amantea}
\IEEEauthorblockA{\textit{Computer Science Department, CIRSFID, SaToSS} \\
\textit{Università degli Studi di Torino, Alma Mater Studiorum, Université du Luxembourg}\\
Turin - Bologna, Italy, Luxembourg\\
amantea@di.unito.it}
\and
\IEEEauthorblockN{2\textsuperscript{nd} Livio Robaldo}
\IEEEauthorblockA{\textit{Legal Innovation Lab Wales} \\
\textit{University of Swansea}\\
Swansea, Wales (UK) \\
livio.robaldo@swansea.ac.uk}
\and
\IEEEauthorblockN{3\textsuperscript{rd} Emilio Sulis}
\IEEEauthorblockA{\textit{Computer Science Department} \\
\textit{Università degli Studi di Torino}\\
Turin, Italy \\
sulis@di.unito.it}
\and
\IEEEauthorblockN{4\textsuperscript{th} Guido Boella}
\IEEEauthorblockA{\textit{Computer Science Department} \\
\textit{Università degli Studi di Torino}\\
Turin, Italy \\
boella@di.unito.it}
\and
\IEEEauthorblockN{5\textsuperscript{th} Guido Governatori}
\IEEEauthorblockA{\textit{Data 61} \\
\textit{CSIRO}\\
Queensland, Australia \\
guido.governatori@data61.csiro.au}
}

\maketitle

\begin{abstract}
 One of the main issues of every business process is to be compliant with legal rules. This work presents a methodology to check in a semi-automated way the regulatory compliance of a business process. We analyse an e-Health hospital service in particular: the Hospital at Home (HaH) service. The paper shows, at first, the analysis of the hospital business using the Business Process Management and Notation (BPMN) standard language, then, the formalization in Defeasible Deontic Logic (DDL) of some rules of the European General Data Protection Regulation (GDPR). The aim is to show how to combine a set of tasks of a business with a set of rules to be compliant with, using a tool. 
\end{abstract}


\section{Introduction}

\noindent The necessity of satisfying regulations or laws forces business organizations to redesign their internal processes, in the context of change management \cite{Hayes:2018}. In the field of Business Process Management (BPM) research, one of the main topic concerns regulatory or Compliance Management (CM), i.e. the analysis of compliance to norms \cite{Dumas-etal:2018}\cite{vanderAalst:2013}.

The increasing pressure from regulatory authorities to organizations led to the development and application of Compliance Management Frameworks (CMFs). In this context, CM can be addressed at the operational level by focusing on business processes, intended as the set of activities accomplishing a specific organizational goal. 
Business process analysis usually introduces performance objectives to be considered in addition to constraints imposed by external pressures (e.g., regulatory issues). The investigation of undesirable events and norm violations adopted traditional techniques, e.g. root cause analysis (commonly used in manufacturing processes to improve performance). More recently, CMFs explore the relationship between the formal representation of a process model and the relevant regulations. There are many different adoptable CM strategies consisting in approaches to automate the checking whether a business process complies with the actual regulation \cite{Governatori-Sadiq:2009}. The goal is to ensure that such approaches properly model business processes as well as norms. Moreover, in the past decades many CM approaches in the context of digitization to automatize business processes have been proposed \cite{Sackmann-etal:2018}. We describe here a CM approach to support regulatory compliance for healthcare business processes based on a compliance-by-design methodology \cite{Governatori-Sadiq:2009} and using a business process compliance checker called Regorous \cite{Governatori:2015}. Regorous implements a propositional logic-based approach of Defeasible Deontic Logic (DDL) \cite{Nute:1997} to automate CM. 

This work is based on a Regional project with European funds called  ``CANP"\footnote{CANP  ``La CAsa Nel Parco": http://casanelparco-project.it/}. The goal of the project is using  Artificial Intelligence and telemedicine solutions to enhance e-Health procedures within the  ``Città della Salute e della Scienza", one of the biggest hospitals in Europe \cite{Sulis-etal:2019}. The application of innovative telemedicine technologies supports the care of elderly patients in the context of a Hospital at Home (HaH). The use of communication systems in the remote management of the patient could improve treatment outcomes, increase access to care, and reduce health costs \cite{Caplan-etal:2012}.

In the paper, we will exemplify our approach by showing how some General Data Protection Regulation (GDPR) provisions to safeguard the personal data of the patients are enforced on the BPM representing the HaH process. However, it will be evident that the approach is general enough to handle any kind of legal constraint involved in BPM procedures.

\section{Background and Related work}

\noindent Legal risks play a crucial roles in business activities, as non-compliant processes may incur heavy sanctions, in particular in the healthcare domain. On the other hand, compliance checking is known to be a difficult task due to the fact that norms are constantly evolving and new reorganizations must be often implemented with the introduction of new procedures 
\cite{Hayes:2018}. For this reason, legal technology
is experiencing growth in activity, also at the industrial level \cite{Ajani-etal:2016}, \cite{Boella-etal:2016}.

Researching the interplay between BPM and CM approaches can lead to the quick identification of processes that need restructuring because they are no longer compliant when regulations are introduced or updated \cite{Racz-etal:2010}.

It is possible to find several studies on compliance with laws, rules or regulations in the case of processes related to patient health \cite{Buddle-etal:2005}, \cite{Amantea-etal:2018}. Compliance in healthcare considers the conformity of care processes with laws, regulations and standards related to patient safety, privacy of patient information and administrative practices.
Ultimately, health compliance is about providing safe and high quality patient care. Healthcare organizations are also required to comply with strict standards, regulations and laws at regional and state level. 

These innovations may require the application of new regulations, such as the GDPR, without forgetting that the health sector is full of strictly health regulations in constant evolution.

\subsection{Business process compliance and logic}

\noindent A business process model is a self-contained, temporal and logical order in which a set of activities are expected to be executed. Typically, a process model describes what needs to be done and when (control flow), who is going to do what (resources), and on what it is working on (data). 

Business Process Management (BPM) is a discipline standardly used to model and analyse business processes, being Business Process Model and Notations (BPMN) the formal language used therein \cite{Allweyer:2016}.

Change management is one of the central issues in BPM research \cite{Abo-Hamad-Arisha:2013}, \cite{Sulis-DiLeva:2018}, \cite{DiLeva-etal:2020}. In the context of healthcare studies, change management is usually handled by creating visual models of processes (i.e., process map or flowchart) and implement them in BPMN \cite{Amantea-etal:2020}, \cite{Sulis2019risk},\cite{DiLeva:2017}, \cite{Muller-Rogge-Solti:2011}. These diagrams depict the sequence of healthcare activities and various crossroads (gateways), which lead to different routes depending on choices made \cite{DiLeva:2017}.

In order to check compliance on business processes, the formal representation of a process and the formal representation of the relevant regulations need to be related \cite{Governatori-etal:2006}.

In the past decades many approaches to automatize business process compliance have been proposed \cite{Becker-etal:2012}, \cite{Ly-etal:2015}. 
However, a challenging research topic is the possibility of modeling standards in a conceptually valid, detailed and exhaustive way that can be used in practice for companies and, at the same time, have the ability to be used generically for any type of standard also taking into account the regulatory environment as a whole \cite{Governatori:2015}.
This shifts the focus to the formalisms used to model regulations.

Linear Temporal Logic and Event Calculus have been used in several frameworks \cite{Awad-etal:2008}. However, it has been shown that when norms are formalized in Linear Temporal Logic the evaluation whether a process is compliant produces results that are not compatible with the intuitive and most natural legal interpretation \cite{Hashmi-etal:2014}, \cite{Governatori:2015a}. Furthermore, it was argued that, while such logics can properly model norms such formalizations would be completely useless from a process compliance point of view insofar they would require an external oracle to identify the compliant executions of the process.

Defeasible Deontic Logic (DDL) \cite{Governatori:2018} has been proposed to overcome these limitations. This is an extension of standard Defeasible Logic with deontic operators, and the operators for compensatory obligation \cite{Governatori-Rotolo:2006}. According to the formalization proposed in \cite{Antoniou-etal:2001}, Defeasible Logic is a constructive logic with its proof theory and inference condition as its core. The logic exploits both positive proofs, a conclusion has been constructively prove using the given rules and inference conditions (also called proof conditions), and negative proofs: showing a constructive and systematic failure of reaching particular conclusions, or in other terms, constructive refutations. The logic uses a simple language, that proved successful in many application area, due to the scalability and constructiveness of the logic. These elements are extremely important for normative reasoning, where an answer to a verdict is often nor enough, and full traceability is needed.

\subsection{Modelling norms in DDL}

\noindent Norms are usually modelled as if-then rules representing the conditions under which norms are applicable and the normative effects these produce when applied. In a compliance perspective, the normative effects of importance are the deontic effects (also called normative positions). The basic deontic effects are: permission, obligation, and 
prohibitions. Thus, an if-then rule in the form ``$a$$\rightarrow$$b$'' that represents a norm reads
``if $a$ holds, then $b$ is permitted/obligatory/prohibited''.
However, it is standardly assumed that prohibitions are indeed obligations; something is prohibited if and only if there is obligation to its contrary \cite{Pattaro-etal:2005}. In this paper, for the sake of simplicity, we assume the term ``obligation'' also encompasses prohibitions.

Obligations  are constraints that limit the space of action of processes. If the antecedent of an if-then rule representing a obligation holds in the context and the bearer of an obligation does not comply with the consequent, then s/he violates the 
obligation. However, a violation does not imply an inconsistency within a process with the consequent termination of or impossibility to continue the business process.
 
Violations can be compensated, and processes with compensated violations are still compliant \cite{Governatori-Sadiq:2009} \cite{Governatori-Milosevic:2005}. E.g., contracts typically contain compensatory clauses specifying penalties and other sanctions triggered by breaches of other contract clauses \cite{Governatori:2005}. Not all violations are compensable; if a process contain uncompensated violations, then it is not compliant.
 
On the other hand, permissions cannot be violated. They can be used (indirectly) to determine that there are no obligations or prohibitions to the contrary, or to derive other deontic effects. Permissions can be 
explicitly encoded in the legislation (positive permissions) or they can be derived, e.g., by observing that there is nothing in the code that forbids them (negative permissions) (cf. \cite{Robaldo-Sun:2017}). 

Instead, compliance usually means to identify whether a process violates or not a set of obligations. Thus, the first step is to determine whether and when an obligation is in force. In other words, an important aspect of the study of obligations is to understand the lifespan of an obligation and its implications on the activities carried out in a process. 

A first broad classification is between punctual obligations and persistent obligations, the former holding on a particular time point, the latter holding over a time interval. In our research, we were only concerned with persistent obligations.

The latter are further sub-classified into maintenance obligations and achievement obligations, the former being in force for all instants in the interval, the latter only until they are achieved/satisfied. Finally, achievement obligations are further sub-classified into preemptive and non-preemptive obligations and between perdurant and non-perdurant obligations. Preemptive obligations are those that could be fulfilled even before the obligation is actually in force. On the other hand, perdurant obligations are those that are still in force even if/after they have been violated.

Regorous, the propositional DDL system we used in our research, formalize the different subclasses of persistent obligations as follows, where ``p'' is a propositional symbol: 
\vspace{3pt}
\begin{itemize}
    \item ``[OM]p'': there is a maintenance obligation for p.
    \item ``[OAPP]p'': there is an achievement preemptive and perdurant obligation for p.
    \item ``[OAPNP]p'': there is an achievement preemptive and non-perdurant obligation for p.
    \item ``[OANPP]p'': there is an achievement non preemptive and perdurant obligation for p.
    \item ``[OANPNP]p'': there is an achievement non preemptive and non-perdurant obligation for p.
\end{itemize}
\vspace{3pt}

On the other hand, positive permissions are represented as ``[P]p'', meaning that ``p'' is permitted. As mentioned above, also positive permissions are involved in the compliance checking process because positive permissions entail that there are no obligations or prohibitions to the contrary or other deontic effects. Furthermore, we may further restrict the process by imposing that it is permitted only what it is explicitly asserted by positive permissions. However, these 
entailments are not relevant here as the case study used in this paper do not encompass deontic reasoning on permissions.

Regorous is implemented as a plug-in of Eclipse. The BPMN is externally built and then uploaded in the platform together with a set of rules in Regorous XML format. Subsequently, in each task of the process it is possible to specify which terms of the vocabulary are true or false via special Eclipse windows provided by the plug-in. Both the BPMN file and the XML file for Regorous are available on GitHub\footnote{\url{https://github.com/liviorobaldo/TrECECpaperAttachments}}.

\vspace{3pt}
\section{Case study: Hospital at Home (HaH) process}

\noindent For more than 30 years, the Hospital ``Città della Salute e della Scienza'' of Turin has operated the Hospital at Home (HaH) service. In this healthcase service, the patient in acute disease (but who do not require equipment with high technological complexity and intensive or invasive monitoring), is taking in charge from the hospital. Specifically, the hospital provides for the organization of care but in the patient's house.

This service is made by two main processes, i.e. the acceptance and the real process of care. We modelled, implemented, and tested both processes, but, for space constraints, this paper will only focus on the acceptance process \cite{amantea2020pm}. The BPMN describing the acceptance process is shown in Figure \ref{RegorousForComplianceChecking} and it is available on GitHub. 

The requests are made by the general medical doctor of emergency or regular departments (generator ``Request HaH''). Each case refers to some guideline to understand if the patient has some characteristics to be taken in charge to this type of hospitalization (tasks ``Make preliminary analysis'', ``Talk to doctor'', ``Talk to patient'', ``Talk to caregiver''). During this meetings, the case manager explains to the patient and to the family the characteristics, organization, and requirements of the service. On the other hand, she collects information out of clinical and functional evaluations as well as cognitive aspects. 

At the end of this evaluation process, the CM decides if either the patient is suitable for an HaH service or if another healthcare process would be more appropriate for his case (``Evaluate alternative route''). In the latter case, the patient will start a new separate healthcare process.

On the other hand, in case the HaH service is selected, the patient or his caregiver need to express his will to access this service or not. From now onward, the process is specifically dealing with the administrative/legal constraints of the HaH service, including those connected with the processing of the patient's personal data. The patient or his caregiver will sign some documents (tasks ``Sign policy of admission'' and ``Compile emergency report'') including, the most important in our case, the informed consent (task ``Fill out the nurse form + Pick up informed consent''), in which the patient or his caregiver will have to provide consent for the processing of the patient's personal data, as required by the GDPR.

After the informed consent is collected, the patient will be registered in the system and taken in charge from the HaH service (tasks ``Make HaH prescription'', ``Make taking in load'', ``Transfer of power''). Other separate healthcare processes will be subsequently carried out in the future in order to implement the HaH service at the patient's home.

\subsection{Encoding GDPR norms in Regorous}

As point out above, Regorous allow to specify permissions and (different) types of obligations holding on propositional symbols ``p''. ``p'' is called a “term” in Regorous terminology. 

Regorous lists all terms used in a set of formalization, together with their description, in a special XML tag 
{\tt <Vocabulary>}. Below, we will show how we modelled Article 6, paragraph 1, point 1 of the GDPR\footnote{\url{https://eur-lex.europa.eu/eli/reg/2016/679/oj}}, which specifies one condition for lawfulness of processing (and indeed the main one): processing of personal data is lawful if ``the data subject has given consent to the processing of his or her personal data for one or more specific purposes;''.

In order to model this norm, we need the concept of personal data processing and the one of consent. These corresponds to two propositional symbols, i.e., in Regorous, to the following two terms that we add to the vocabulary:

\vspace{3pt}
\begin{alltt}
  <Vocabulary>
     <Term atom="Proc" 
       description="Processing: means 
       any operation or set of operations 
       which is performed on personal data
       ..."/>
     <Term atom="GiveConsent"
       description="Consent given by
       the data subject means any freely
       given, specific, informed and
       unambiguous indication ..."/>
  </Vocabulary>
\end{alltt}
\vspace{3pt}

Given the two terms in the vocabulary, we decided to implement the GDPR norms connected with the lawfulness of processing by forbidding processing of personal via a general maintenance obligation, unless a more specific (and stronger) if-then rule, based on given consent, permits it.

The maintenance obligation is formalized as follows:

\vspace{3pt}
\begin{alltt}
 <Rule xmlns:xsi="..." ruleLabel="Art.6.0"
                  xsi:type="DflRuleType">
   <ControlObjective>Personal data 
      processing is prohibited.
   </ControlObjective>
   <FormalRepresentation>
      =>[OM]-Proc
   </FormalRepresentation> 
 </Rule>
\end{alltt}
\vspace{3pt}

In Regorous, negation and logical entailment are represented via the symbols ``{\tt -}'' and ``{\tt =>}''. Therefore, the antecedent of the if-then rule ``{\tt =>[OM]-Proc}'' is empty, meaning that the consequent always holds, i.e., that the negation of the term ``{\tt Proc}'' is obligatory for all over the process.

On the other hand, the permission based on given consent is formalized as follows:

\vspace{3pt}
\begin{alltt}
<Rule xmlns:xsi="..." ruleLabel="Art.6.1a"
                  xsi:type="DflRuleType">
   <ControlObjective>Processing shall be 
   lawful if the data subject has given 
   consent to the processing of his or 
   her personal data for one or more 
   specific purposes.</ControlObjective>
   <FormalRepresentation>
      GiveConsent=>[P]Proc
   </FormalRepresentation> 
</Rule>
\end{alltt}
\vspace{3pt}

In this case, the antecedent of the if-then rule is not empty: the rule will triggers only when the term 
{\tt GiveConsent} is asserted. In our BPMN, this will be done in the task ``Fill out the nurse form + Pick up informed consent''.

Of course, the two formulas cannot hold together as the first entails that the processing is prohibited while the latter entails that it is permitted. In order to solve these conflicts, DDL implements overriding relations between norms. In our example, the permission {\tt [P]Proc} must override the obligation {\tt [OM]-Proc} for the business process to be compliant with the GDPR.

In Regorous, DDL overriding relations are implemented as ``superiority relations'', encoded via the homonym tag, in which the ``superiorRuleLabel'' overrides the ``inferiorRuleLabel''. Therefore, in the example under consideration we have:

\begin{alltt}
   <SuperiorityRelation 
          superiorRuleLabel="Art.6.1a" 
          inferiorRuleLabel="Art.6.0"/>
\end{alltt}

By asserting these superiority relations, Regorous is able to infer that if both the terms {\tt Proc} and {\tt GiveConsent} holds, the process is compliant. 

The assertion of {\tt GiveConsent} does not only entail that processing of personal data is permitted. According to Article 7(1) of the GDPR, ``Where processing is based on consent, the controller shall be able to demonstrate that the data subject has consented to processing of his or her personal data.''. Therefore, the assertion of 
{\tt GiveConsent} also entails another maintenance obligation for the hospital, formalized in Regorous as follows:

\vspace{3pt}
\begin{alltt}
<Rule xmlns:xsi="..." ruleLabel="Art.7.1"
                  xsi:type="DflRuleType">
 <ControlObjective>If processing is
   based on consent, the controller shall
   be able to demonstrate the consent.
 </ControlObjective>
 <FormalRepresentation>
     GiveConsent=>[OM]DemonstrateConsent
 </FormalRepresentation> 
</Rule>
\end{alltt}
\vspace{3pt}

The obligation is again a maintenance obligation, rather than an achievement one, because the GDPR requires the controller (the hospital, in our case) to store and provide evidence the given consent {\it at any time}, upon request of the data protection authority or any other appointed auditor. In the HaH process, the personal data record for the patient, which includes the evidence for his given consent, is prepared  and stored in the hospital digital archive in the task ``Make taking in load'', together with all other administrative details referring to the HaH service that will be put in place. At the same time, the term {\tt DemonstrateConsent} will be asserted in Regorous. The term will remain true until the record is kept in the hospital archive, so that Regorous will be always able to assess compliance with {\tt [OM]DemonstrateConsent} that, being a maintainance obligation, must be satisfied in every instant of the execution.

Note that in our use case, we are focusing only on Article 6.1(a) of the GDPR. However, given consent is only one of the legal bases that allows for processing of personal data, the one that is used by the HaH acceptance process. On the other hand, our formalization includes other legal bases, specified in other norms of the GDPR, and used in other healthcare processes within the hospital.

For instance, Article 6.1(b) of the GDPR permits processing if it is ``necessary for the performance of a contract to which the data subject is party or in order to take steps at the request of the data subject prior to entering into a contract'' while Article 6.1(d) of the GDPR permits processing if it is ``necessary in order to protect the vital interests of the data subject or of another natural person.''.

Article 6.1(b) and Article 6.1(d) provides then other two sufficient conditions to permit processing of personal data, so that they are respectively formalized in the if-then rules ``{\tt Contract=>[P]Proc}'' and ``{\tt VI=>[P]Proc}''. Two corresponding superiority relations are also added to the Regorous knowledge base in order to allow the overriding of {\tt [OM]-Proc} also in these two cases.

The fact that these two sufficient conditions are not used in the BPMN under consideration highlights the fact that the Regorous file is not 1:1 associated with the HaH acceptance process. In other words, the formalization of the legislation is kept independent from the modeling of the healthcare processes and shared among multiple ones.

In case new legislation will enter into force, or, alternatively, the existing one is updated, only the {\tt <Rule>}(s) will need to be modified at first. Regorous will be then able to identify which business processes are affected by the changes: those that are no longer compliant with the new legal setting.

\section{Executing Regorous on the business process}

Given a set of well-formed rules and superiority relations encoded in the XML format briefly seen above, Regorous allows to check whether a Business Process in the BPMN standard is compliant with them. 

Regorous is implemented as a plug-in of Eclipse. 
After the creation or the import of BPMN process it is possible to upload of the file or the files with the set of rules in Regorous XML format. Afterwards, in each activity of the process it is possible to specify the resources used and for each of them the detailed tasks that are included in the activity. Both the resourses and the actions have to be express using the name of the vocabulary  terms. 

In this way, via a special Eclipse windows provided by the plug-in, it is possible to see if the process is compliant, not compliant, or also maybe compliant (with some warning). 

\begin{figure*}
%\subfloat{\includegraphics[width=1\textwidth]{Reg.png}} 
\includegraphics[width=1\textwidth]{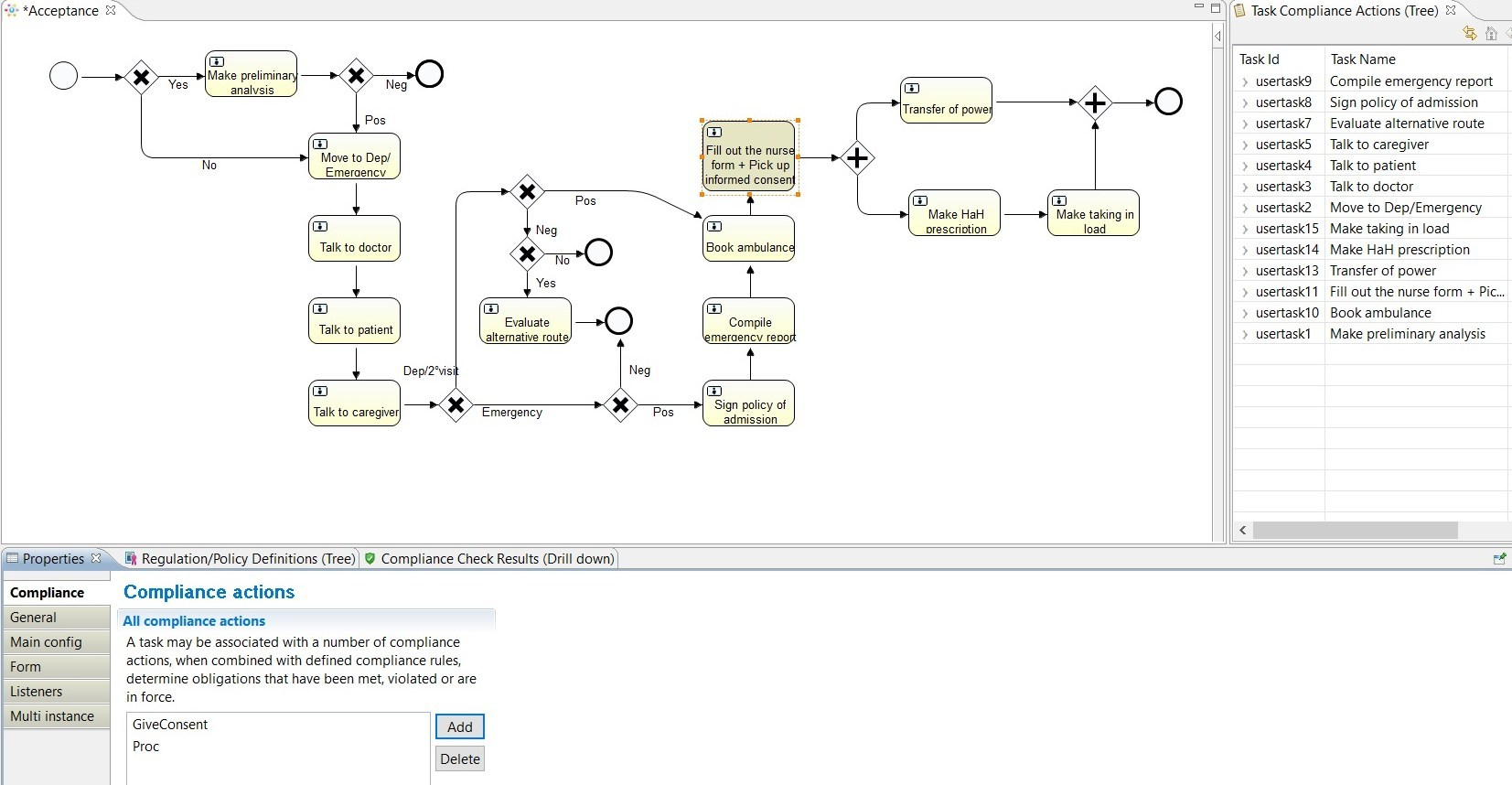}
\caption{Regorous screenshot example for compliance checking}
\label{RegorousForComplianceChecking}
\end{figure*}

Just to show a practical example, Figure \ref{RegorousForComplianceChecking} shows how Regorous performs compliance checking on the HaH process. The BPMN file is uploaded in Eclipse together with the ruleset formalizing the GDPR norms in Regorous XML format.

Precisely, as shown on the bottom of Figure  \ref{RegorousForComplianceChecking}, the plug-in includes special tabs that allow to specify, for each task, the values of the terms. For instance, by specifying “GiveConsent” and “Proc” in the task “Fill out the nurse form + Pick up informed consent”, Regorous infers that the process is compliant with the ruleset, as the superiority relation seen above will make the processing of personal data permitted. Conversely, by specifying the single action “Proc”, Regorous infers that the process is not compliant with the ruleset because the rule with ruleLabel=“Art.6.0” asserts the processing of personal data as prohibited and, contrary to the previous case, that prohibition is not overridden by a stronger permission.

After setting the activities, the compilation is launched. At this point it is important to underline some things. The more important is that the check follows the process flow. Therefore, it starts from the generator (the rod on the left) and runs the various flows for each branch (gateway) of the process and till all the various possible ends of the process. In this way, the first import thing is that all possible individual paths are checked.

Moreover, each path is not run once, but several times. Here the rules of superiority come into play and therefore it is possible to verify even if some activities are generically controlled sooner or later within the process, if they are for all possible branches and if they are (for each branch) in the sequence established by law.
    
In fact, a rule can, for example:
\begin{itemize}
    \item generically establish that an action must be carried out. In this case, by running the run more than once it is possible to check whether the action is carried out sooner or later in an activity of the process, but also if this activity is not only present, but is present for all possible branches of the process. For example, it is possible that one requirement may be overridden if another occurs. For example, in our case the rule is that personal data cannot be processed, this is the primary rule. But then the rule allows exceptions, for example if consent is given. Of course, we don't know from the beginning, by law, if and when (in what activity) consent will be given.Therefore, the primary rule will apply until the requirement contained in the superior rule of consent arrives. If there were no rules of superiority, the primary requirement that personal data cannot be processed would continue to apply. Furthermore, if for example the consent were revoked, this would be a rule superior to the "give consent" rule which would therefore annul the consent previously given. In this case, however, it should be noted that it is as if the consent had not been obtained from the beginning, therefore, if, for example, other branches of the process are in progress that are in the meantime being carried out in parallel, they may no longer be compliant. If the Regorous control stops at the first run, without possibly having some subsequent steps in memory, it could provide incorrect output. It would be enough for the consent to be given before by a gateway which provides for two parallel activities / paths and then to be revoked in only one of two branches (for example, the patient would probably communicate the revocation to the doctor or nurse, but not to any laboratory).
    \item provide that one activity must be done following or before another. For example, in our case the consent must be signed after having received the information and before having the treatment. Therefore, it is not only important that the activities of speaking, signing, being taken in charge exist, but it is important that, in any branch of the process, if there is the signing activity, it is preceded by having had the information and if a patient is registered is preceded by having signed consent.
\end{itemize}

In this way, at the end of the execution, an output will appear in which it will be explained if:
\begin{itemize}
    \item The process is compliant (in green). It means that the whole process, therefore all the paths that form the process are completely compliant. In this case there will be no other information.
    \item The process is not compliant (in red). It means that there is one or more lack of mandatory controls or activities. In this case they will also be explained what they are. They will be indicated with the names used in the Vocabulary and the missing control indicated in the formalized law.
    \item The process may not be compliant (in orange). This is a warning. It means that, for example, there could be the mandatory activities but in some branch they may not be in the right sequence, therefore it is indicated where to carry out the check.
\end{itemize}

This is obviously an example case to show the methodology. however, it is important to underline two other aspects:
\begin{itemize}
    \item Once a law has been formalized, the file remains saved, so in the event of partial changes by the legislator it will be sufficient to modify only the new part.
    \item More than one file can be uploaded, one per regulation. Probably any company must be subject to more than one law (for example that of detail, the national one, the European one. Or in different fields, such as that of the sector, that on employment contracts, that on security, and that on privacy). Being able to enter and therefore check compliance on more than one regulation at the same time, it is also possible that Regorous (in orange) signals steps that could be compliant for a regulation but not fully correct for a second.
\end{itemize}

\section{CONCLUSION AND FUTURE WORKS}

The proposed pipeline addresses a specific risk management application in a selected healthcare process. However, of course much further work is needed to formalize all the laws, regulations, and guidelines involved in all healthcare processes, in order to have a full exhaustive analysis on how legal compliance is handled in medical procedures.
The research presented here requires manual formalization of the regulations, a task that we aim at semi-automatizing in our future work by following the methodology presented in \cite{Bartolini-etal:2016}. On the other hand, following \cite{Robaldo:2021}, we aim at integrating automatic reasoning on the formalized  norms fit to further assist and guide the creation of large repositories or rules, to be possibly shared among multiple business processes.

In addition, the aim is to combine this compliance checking methodology with a framework that also includes a change in the flow of activity in a context of re-organization and optimization of processes. Maintaining the already formalized norms as a background, the purpose is to obtain a methodology able to balance the managerial aspect with that of regulatory compliance. 

In summary, this methodology, which is based on processes, uses the modeling of them in the standard BPMN language and which allows you to export the process or import one already created previously via a .bpm file, has a particular advantage. The .bpm extension is quite common for business programs, therefore, the file containing the process could be used by other tools, such as business simulators or value management platforms, to consider other aspects of the company. 
In this way, it would be easier to balance the purely organizational aspects of the business (resources, production times and costs or qualitative values of the service) with the aspect of regulatory compliance.

This aspect is becoming more and more important as the legislative evolution is in continuous and increasingly rapid evolution and companies must adapt and change more and more rapidly to remain competitive on the market.

On the other side, it is then important to note that the BPMN and the XML configuration file of Regorous will always need human intervention to be modified. In other words, Artificial Intelligence is only conceived as a tool to only assist and monitor humans in their work activities, but not to take decisions in their place.

The different modules required by the Artificial Intelligence can be prepared and updated by hospital departments with different expertise and authority, thus enhancing efficiency and communication. For instance, the Regorous configuration files can be prepared and maintained by the legal department of the hospital, while the BPMNs by the department concerned with logistics. Regorous will then provides an implicit and efficient communication channel between the two departments, so that each of them will be able to focus on its own competence field while minimizing interactions with the other departments (arrange specific inter-department meetings, etc.)

In the context of the project CANP, the authors have also started working on two directions of improvement for the framework presented here, which will become the main object of future works:

\begin{itemize}
    \item The first is to try other deontic logics, for instance reified I/O logic \cite{Robaldo-etal:2020} that, being first-order, provides an higher degree of expressivity as well as a direct interface with OWL \cite{OWL:2012}, the W3C standard for the semantic web.
    \item Secondly, as mentioned above we are carrying out researches to verify whether it is possible to semi-automate the manual encoding step of the law via NLP \cite{Sulis-etal:2021,amantea-model:2019}. This is the most critical point, as it is the most crucial for companies but, at the same time, the most complex to automate as it is necessary to take into account the interpretation of the laws.
\end{itemize}

% However, in the context of the project “<blind-reference>”, the authors are also currently working on the development of a methodology to automatize or semi-automatize formalization of laws, that combines Defeasible Deontic Logic with NLP technologies, in order to make the whole process faster, simpler, and accessible to users who have little or no competence in law or in logical formalizations.

\section*{Acknowledgment}

The current research was conducted in the project ``CANP - CAsa Nel Parco'' of Regione Piemonte funded by POR FESR PIEMONTE 2014-2020. We are grateful for the collaboration of the ``City of Health and Science'' of Torino (Italy). We are also grateful to the partners involved in the project. Livio Robaldo has been supported by the Legal Innovation Lab Wales
operation within Swansea University’s HRC School of Law; the operation has been part-funded by the European Regional Development Fund through the Welsh Government.

\vspace{12pt}

\end{document}